\documentclass[journal]{IEEEtran}

\usepackage{cite}
\usepackage{amsmath,amssymb,amsfonts}
\usepackage{algorithmic}
\usepackage{multirow}
\usepackage{graphicx}
\usepackage{textcomp}
\usepackage{booktabs}
\usepackage{makecell}
\usepackage{float}
\usepackage{amssymb}
\usepackage{color}
\usepackage{amsmath}

\begin{document}
\title{AGMB-Transformer: Anatomy-Guided Multi-Branch Transformer Network for Automated Evaluation of Root Canal Therapy}
\author{Yunxiang Li*, Guodong Zeng*, Yifan Zhang, Jun Wang, Qun Jin, \IEEEmembership{Senior Member, IEEE}, Lingling Sun, Qianni Zhang, Qisi Lian, Guiping Qian, Neng Xia, Ruizi Peng, Kai Tang, Shuai Wang$^{\dagger}$, Yaqi Wang$^{\dagger}$

\thanks{This research is supported in part by the National Natural Science Foundation of China (No. 61827806), National Key Research and Development Program of China (No. 2019YFC0118404) and Public Projects of Zhejiang Province (No. LGG20F020001).}
\thanks{* Co-first author. }
\thanks{$\dagger$ Corresponding author.}
\thanks{Yunxiang Li, Lingling Sun, Neng Xia, Ruizi Peng, Kai Tang are with Microelectronics CAD Center, Hangzhou Dianzi University, Hangzhou, China (e-mail:li1124325213@hdu.edu.cn, sunll@hdu.edu.cn, amoreyo@hdu.edu.cn, prince75@hdu.edu.cn, 19061129@hdu.edu.cn) }
\thanks{Guodong Zeng is with sitem Center for Translational Medicine and Biomedical Entrepreneurship, University of Bern, Bern, Switzerland (e-mail: guodong.zeng@sitem.unibe.ch).}
\thanks{Yifan Zhang, Qisi Lian are with National Clinical Research Center for Oral Diseases, West China Hospital of Stomatology, Sichuan University, Chengdu, China (e-mail: zhangyifan@hzyk.com.cn, qisiscu@163.com).}
\thanks{Jun Wang is with School of Biomedical Engineering, Shanghai Jiao Tong University, Shanghai, China (e-mail: wjcy19870122@sjtu.edu.cn).}
\thanks{Qun Jin is with the Department of Human Informatics and Cognitive Sciences, Faculty of Human Sciences, Waseda University, Tokyo, Japan (e-mail: jin@waseda.jp).}
\thanks{Qianni Zhang is with School of Electronic Engineering and Computer Science, Queen Mary University of London, London, UK (e-mail: qianni.zhang@qmul.ac.uk).}
\thanks{Guiping Qian, Yaqi Wang is with the College of Media Engineering, Communication University of Zhejiang, Hangzhou, China (e-mail: qianguiping@163.com, wangyaqi@cuz.edu.cn).}
\thanks{Shuai Wang is with School of Mechanical, Electrical and Information Engineering, Shandong University, Weihai, China (e-mail: shuaiwang@sdu.edu.cn).}
}

\maketitle
\begin{abstract}
Accurate evaluation of the treatment result on X-ray images is a significant and challenging step in root canal therapy since the incorrect interpretation of the therapy results will hamper timely follow-up which is crucial to the patients' treatment outcome. Nowadays, the evaluation is performed in a manual manner, which is time-consuming, subjective, and error-prone. In this paper, we aim to automate this process by leveraging the advances in computer vision and artificial intelligence, to provide an objective and accurate method for root canal therapy result assessment. A novel anatomy-guided multi-branch Transformer (AGMB-Transformer) network is proposed, which first extracts a set of anatomy features and then uses them to guide a multi-branch Transformer network for evaluation. Specifically, we design a polynomial curve fitting segmentation strategy with the help of landmark detection to extract the anatomy features. Moreover, a branch fusion module and a multi-branch structure including our progressive Transformer and Group Multi-Head Self-Attention (GMHSA) are designed to focus on both global and local features for an accurate diagnosis. To facilitate the research, we have collected a large-scale root canal therapy evaluation dataset with 245 root canal therapy X-ray images, and the experiment results show that our AGMB-Transformer can improve the diagnosis accuracy from 57.96\% to 90.20\% compared with the baseline network. The proposed AGMB-Transformer can achieve a highly accurate evaluation of root canal therapy. To our best knowledge, our work is the first to perform automatic root canal therapy evaluation and has important clinical value to reduce the workload of endodontists. 
\end{abstract}


\begin{figure}
\centering
  \includegraphics[width=.5\textwidth]{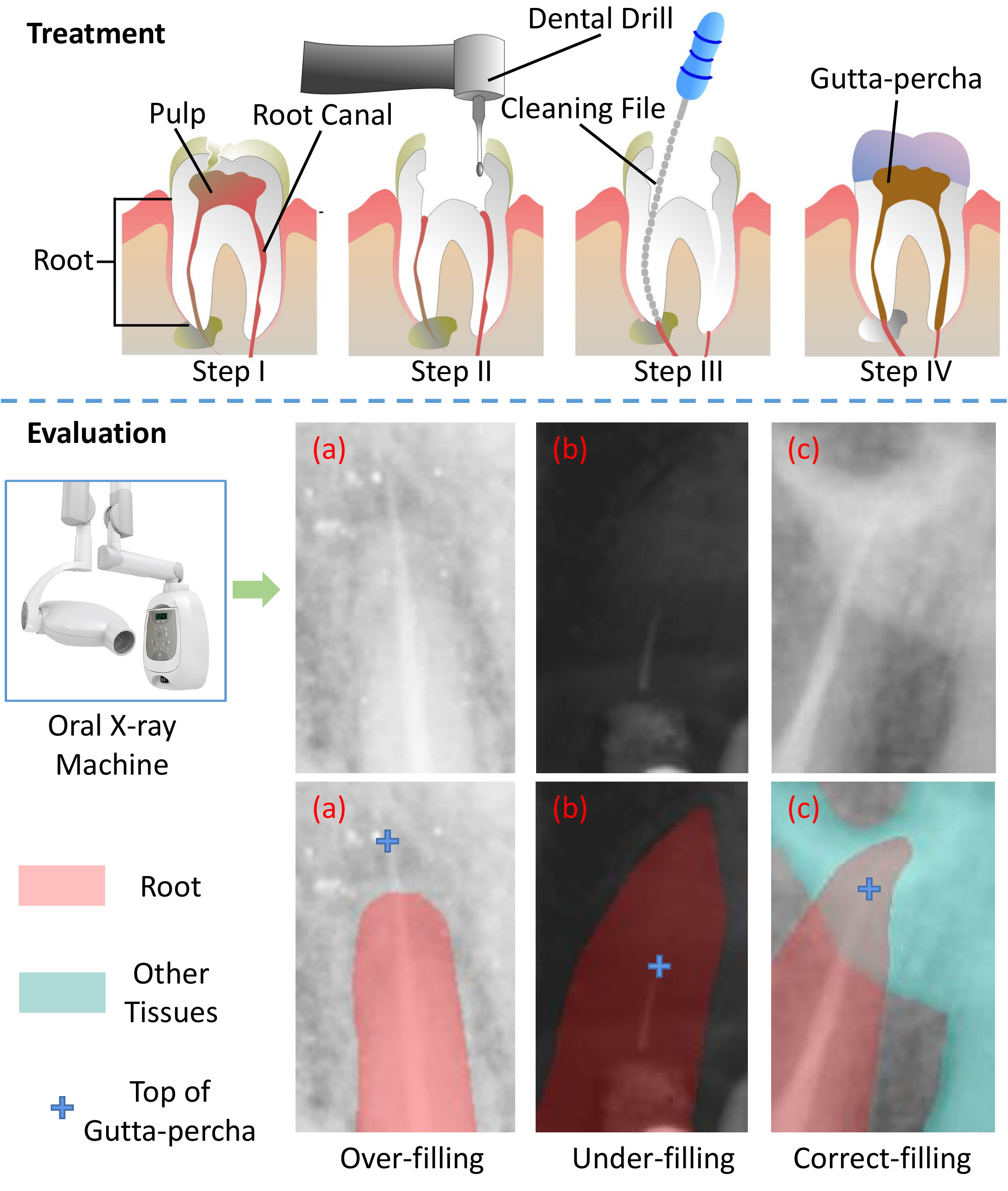}
  \caption{The first part shows the conventional steps of root canal therapy. The second part gives examples of three types of root canal therapy results, and the first row shows original images and the second row shows the corresponding images overlaid with red, blue, and cyan to represent some anatomy features.}
  \label{fig:challenge}
\end{figure}

\section{Introduction}
Severe periodontitis is the sixth-most prevalent health condition, affecting approximately 10\% of people worldwide\cite{peres2019oral}. Root canal therapy is a common method in periodontal disease treatment, but it is easily prone to errors\cite{meirinhos2020prevalence,boucher2002radiographic,lazarski2001epidemiological,kaplan2019dental}, which leads to a significant negative impact on patient outcomes\cite{lin2005procedural,estrela2014characterization,saunders1997technical,petersen2005global}. More precisely, under-filling will lead to the bacterial residue, which can form acute or chronic inflammation, and patients with over-filling may experience complications such as pain or tissue necrosis, even more severe, resulting in neurologic complications such as hyperaesthesia or dysaesthesia\cite{kim2016accidental}. Therefore, correct evaluation in post-therapy is very significant since under-filling and over-filling need to be retreated and remedied in time. Nowadays, the evaluation of a root canal therapy result is relied on the personal empirical assessment of endodontists, to decide it as a case of under-filling, over-filling, or correct-filling\cite{field2004clinical}. Unfortunately, there are many bottlenecks in this manual process. Firstly, the manual evaluation of multiple X-ray images in each case is time-consuming and tedious work. Secondly, the manual evaluation results are easily prone to errors due to the requirement of high-level expertise and are highly subjective due to unavoidable inter-observer variability. Therefore, an automatic and accurate evaluation method for the root canal therapy results is highly desired to improve the diagnosis accuracy and efficiency, and at the same time, reduce the cost associated with human effort and finance.

\begin{figure*}[ht]
  \centering
  \includegraphics[width=1.\textwidth]{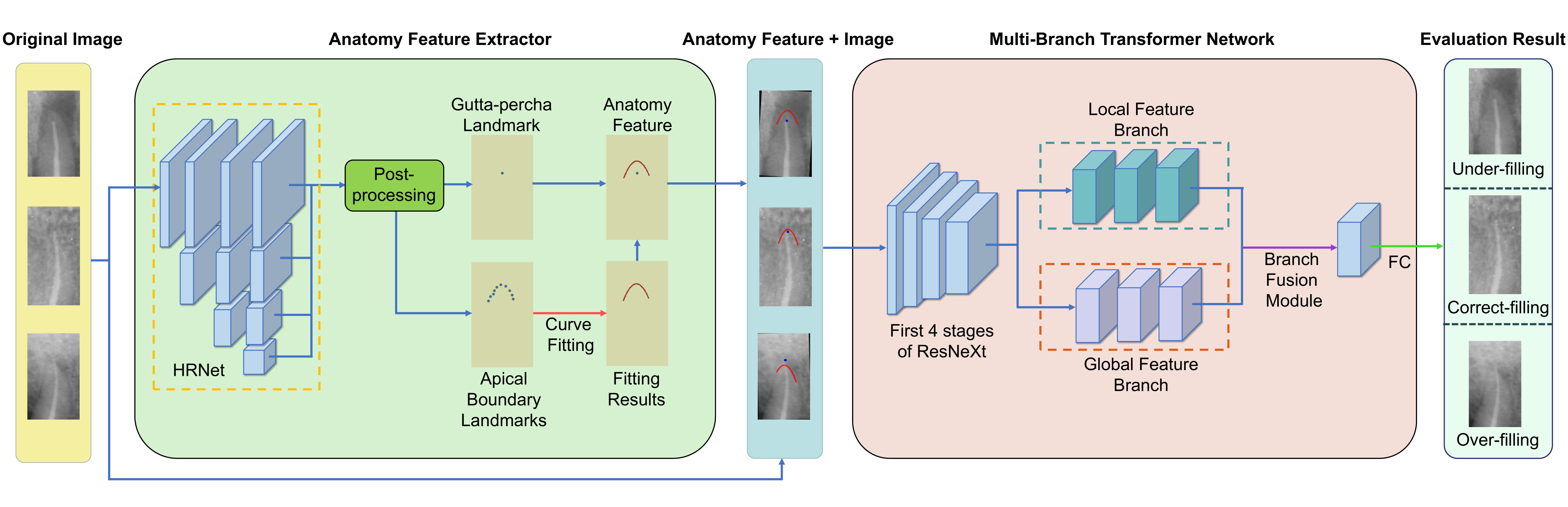}
  \caption{An illustration of the proposed AGMB-Transformer. First, anatomy features are generated by fitting segmentation and landmark detection. Then, anatomy features are combined with the X-ray images as the input for the AGMB-Transformer. In order to better visualize the process, we mark the segmentation results with red lines and the landmarks with blue points. }
  \label{fig:allprocess}
\end{figure*}

However, the evaluation of root canal therapy is a challenging task to computers due to many intrinsic complexities: 1) the apices area boundaries of tooth roots, which are the most important anatomy cues for the evaluation, are difficult to extract using conventional edge detection methods since the boundaries of the teeth in X-ray images are unclear and irregular. An example is shown in \textcolor[rgb]{0,0.541,0.855}{Fig.\ref{fig:challenge} (a)}; 2) the imaging quality of some X-ray images is poor (overexposed or underexposed) that even most experienced endodontists find it hard to make an accurate visual assessment, as shown in \textcolor[rgb]{0,0.541,0.855}{Fig.\ref{fig:challenge} (b)};    
and 3) the tooth root may be obscured by other bones and tissues of the head and cannot depict adequate visual information for the evaluation, as shown in \textcolor[rgb]{0,0.541,0.855}{Fig.\ref{fig:challenge} (c)}.

To address these challenges, we propose an anatomy-guided multi-branch Transformer network that jointly models the global and local visual cues with the assistance of anatomy features to make an accurate evaluation of root canal therapy. The detailed architecture of the proposed network framework is illustrated in \textcolor[rgb]{0,0.541,0.855}{Fig. \ref{fig:allprocess}}. Specifically, in the anatomy feature extractor module, the apical boundary of the tooth root and the top of the gutta-percha in the root canal are encoded as anatomy features and are extracted based on a landmark detection strategy and a polynomial curve fitting approach. Then, the anatomy features, together with the original images, are fed as the input to the proposed multi-branch Transformer network, which considers both global and local features through two branches - group convolution and progressive Transformer. To avoid the curve blocking the information from the original image, the curve covers the original image with the width of one pixel. Finally, a branch fusion module is developed to effectively fuse the features from the two branches and make a joint prediction.

The main contributions of this paper are listed as follows: 

\begin{itemize}
    \item To our best knowledge, this is the first research that realizes an automatic and accurate method for root canal therapy evaluation, which can potentially bring significant benefits for root canal therapy in clinical practice.

    \item To effectively extract the anatomy cues, we propose a novel anatomy feature extractor that is suitable for detecting the fuzzy boundaries in oral X-ray images. It achieves an accurate tooth apices area segmentation through a polynomial curve fitted to detected landmarks.

    \item We design a multi-branch Transformer network, combining the advantages of group convolution and progressive Transformer, where our progressive Transformer achieves multi-scale self-attention and reduces the amount of computation through Group Multi-Head Self-Attention. Our novel network structure solves the intrinsic locality issue of convolution and accomplishes multi-branch structure features fusion.
\end{itemize}

\section{Related Work}
In this paper, we propose a deep learning-based method for the evaluation of the root canal therapy results. Our method entails two main components: an anatomy feature extractor that relies on a segmentation approach, and a root canal therapy evaluation method following a classification approach. This research mainly involves the two fields: medical image classification and segmentation\cite{9286475,8861329,9122550,7523288,9269406}. Thus, the most related works in these two fields are reviewed in this section.

\subsection{Medical Image Segmentation}

In many biomedical image modalities, the important region boundaries are fuzzy despite the best possible image quality, including the root boundaries in oral X-ray images. It is a challenging task to segment the apical area boundaries of tooth roots accurately, while the apical boundaries are the key anatomy features for root canal therapy evaluation.
To solve the problem of tooth root segmentation, Zhao et al.\cite{zhao2020tsasnet} proposed a two-stage attention segmentation network for the tooth segmentation task, following a similar approach to Attention U-Net\cite{oktay2018attention}. Lee et al.\cite{lee2020application} proposed a deep-learning method using a fine-tuned mask R-CNN\cite{he2017mask} algorithm. Koch et al.\cite{koch2019accurate} improved U-Net by using patches as inputs replacing full images. However, these methods do not solve the segmentation problem of fuzzy boundaries, and the performance improvement is mostly incremental. Cheng et al.\cite{cheng2020learning} proposed U-Net+DFM to learn a direction field. It characterizes the directional relationship between pixels and implicitly restricts the shape of the segmentation result. Their method can obtain a significant improvement on the fuzzy boundary segmentation task, but it is still limited by the accuracy of U-Net\cite{ronneberger2015u}. Moreover, an efficient anatomy feature for the evaluation of root canal therapy is the apical area boundary of the tooth, but traditional dental segmentation methods all aim to segment the whole tooth, occupying unnecessary computational resources in the less relevant parts of the tooth. Many current methods are based on the variations of U-Net, and their improvements compared to U-Net are incremental. However, the boundary of the apical region is extremely vague, and there is no model that can accurately segment it to meet the precision requirements of root canal therapy evaluation in the existing segmentation methods.

\subsection{Medical Image Classification}
Correct classification of root canal therapy results is the ultimate goal of this research. Similarly, in many medical image analysis tasks, classification is an important solution to disease diagnosis or grading. A considerable amount of effort has been dedicated to the automatic classification of medical images. Traditional classifiers mostly rely on handcrafted features such as Random Forest (RF)\cite{gray2013random}, Support Vector Machine (SVM)\cite{noble2006support}, and Multi-Layer Perceptron (MLP)\cite{tang2015extreme}. However, it is recognized that hand-crafted features are not descriptive enough for representing the comprehensive structural diversity in most medical images like X-ray, and more effective methods need to be explored.

\subsubsection{Convolutional Neural Networks}
Recent advances in deep learning and computer vision have indicated neural networks are more suitable for automatic feature learning\cite{shen2017deep}. In particular, Convolutional Neural Networks (CNNs) have been widely applied to achieve state-of-the-art results in various computer vision and biomedical image analyses. The mass application of CNNs\cite{9362125,7930382,9316292} reveals their advantage in solving image classification problems and illuminates a promising direction for medical evaluation tasks by using CNNs models to explore inconspicuous local features. The most widely used CNN is ResNet proposed by He et al\cite{he2016deep}. Based on it, Xie et al. developed ResNeXt\cite{xie2017aggregated}, and Hu et al. designed SENet\cite{hu2018squeeze}. ResNeXt and Xception\cite{chollet2017xception} introduces group convolution and SEResNeXt is a commonly used fine-grained classification baseline. Besides, Huang et al.\cite{huang2017densely} proposed DenseNet, which adopts dense connections to avoid the vanishing gradient problem. Unfortunately, the great success of CNNs relies heavily on the amount of training data and on obvious features for classification. Learning high-level semantic information for fine-grained image classification remains a challenging task for classification networks. Moreover, CNNs suffer from an intrinsic limitation that they pay more attention to local information, and thus are not as capable of processing global features.

\subsubsection{Transformers}
To address the limitations of CNNs, a new architecture is necessary to extract global features. Transformer-based architectures employ position encoding, which has recently been proven to be more suitable for many vision applications\cite{han2020survey}.
It was first proposed by Vaswani et al.\cite{vaswani2017attention} for machine translation and has become the state-of-the-art method in most Natural Language Processing (NLP) tasks, where multi-head self-attention (MHSA) is employed as the base structure. In addition, in the process of self-attention exploration, much progress is being made by researchers. Wang et al.\cite{wang2018non} presented a novel type of neural networks that capture long-range dependencies via non-local operations. Within this general framework, Yue et al.\cite{cao2019gcnet} designed a better instantiation named global context network (GCNet), which is lightweight and able to effectively model the global features. Although these network structures can achieve long-range dependencies, Transformer gives a modeling pipeline that does not rely on CNNs to achieve that. Due to the difference between NLP and computer vision, it is difficult to apply transformers directly to image processing tasks. In order to design a transformer structure suitable for image processing, several approaches have been proposed in recent years. Parmar et al.\cite{parmar2018image} applied self-attention for each query pixel only in local neighborhoods rather than globally. Nicolas et al.\cite{carion2020end} presented DETR, a new design for object recognition based on Transformer that significantly improved the performance over Faster R-CNN for large objects. Alexey et al.\cite{dosovitskiy2020image} explored the direct application of Transformer to image recognition. This application is called ViT and interprets images as a series of patches and then processes them using an off-the-shelf Transformer encoder. Due to the lack of translation equivariance and locality, Transformer does not perform well on insufficient datasets. The amount of training data required for ViT is 14-300 million images. The combination of Transformer and convolution is considered an effective scheme to reduce the scale of training data, and thus Aravind et al.\cite{srinivas2021bottleneck} presented BoTNet, a simple but powerful instance segmentation and object detection backbone, where a Bottleneck Transformer is used as a base structure by simply replacing a part of convolutions with MHSA in bottleneck blocks. 


\section{Method}
The proposed anatomy-guided multi-branch Transformer network comprises two stages. The first stage is anatomy feature extraction that works by extracting the apical boundary of the tooth root and the landmark of the gutta-percha in the root canal. In the second stage, the extracted anatomy features together with the X-ray images are fed into a multi-branch Transformer network for deciding the final evaluation outcome.

\begin{figure}[ht]
\centering
  \includegraphics[width=.5\textwidth]{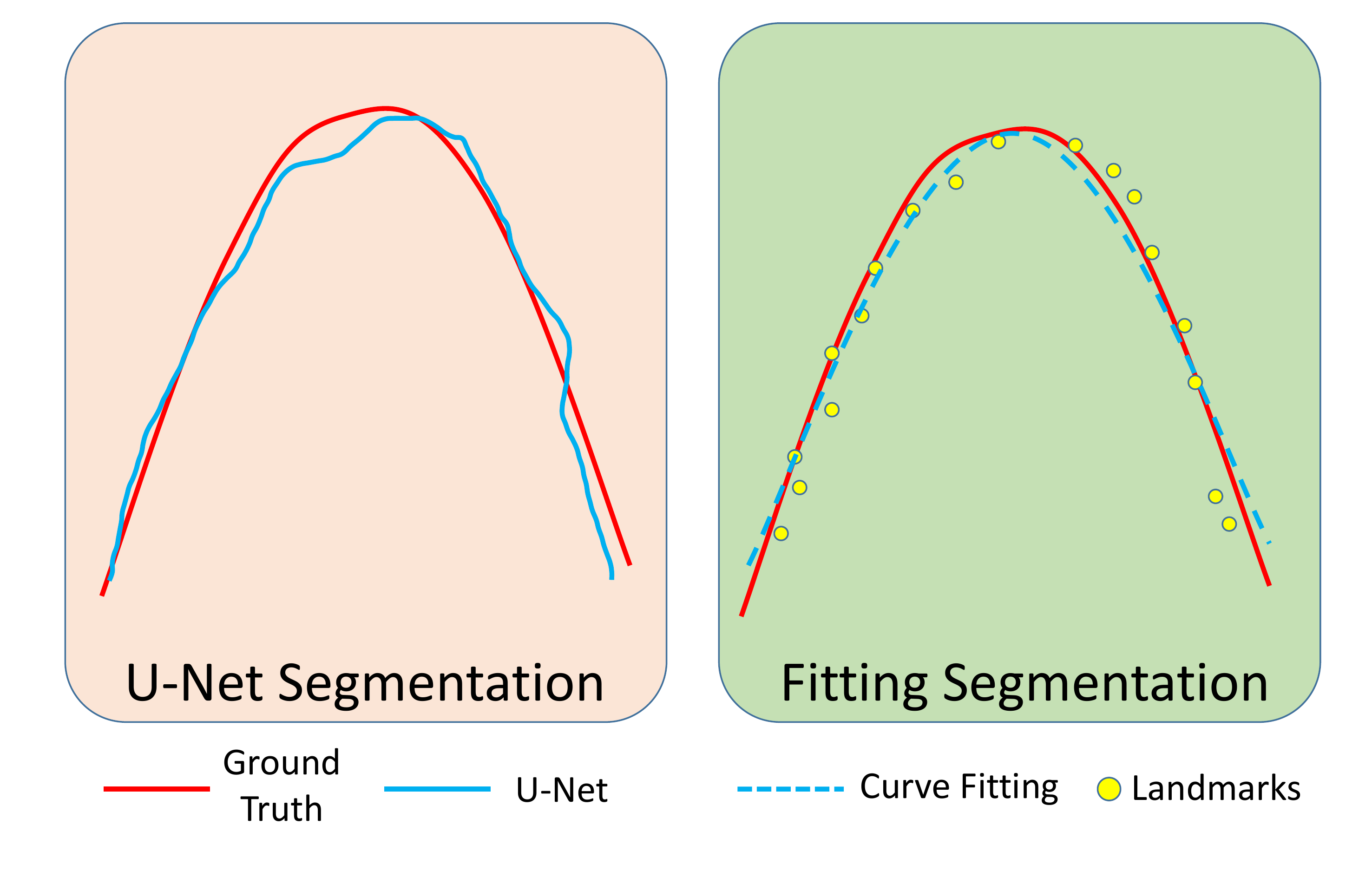}
  \caption{The comparison of traditional segmentation network U-Net and our proposed fitting segmentation method in the segmentation of extremely fuzzy boundaries.}
  \label{fig:osl}
\end{figure}

\subsection{Anatomy Feature Extractor}
In medical imaging, remarkable progress has been made in high-performance classification models based on CNNs. Despite the performance peaks, new advanced neural networks still require large, representative, and high-quality annotated datasets. Unfortunately, both data and annotations are expensive to obtain\cite{tajbakhsh2020embracing,pasupa2016comparison}. One method that can reduce the dependency on large-scale annotated data for training without sacrificing classification performance is to leverage anatomy features as prior knowledge to facilitate the classification model\cite{tran2021segmentation}. The root canal therapy evaluation is a very subjective process, which requires comprehensive consideration of cues such as the shape and location of the tooth apical boundary, the position of gutta-percha filling in the root canal, the X-ray projection angle, etc, with a particular emphasis on the relative position of the apical boundary and the top of the gutta-percha. Consequently, the performance of the classification network could be improved if the apical boundary of the tooth root and the top of the filled gutta-percha were to be considered as anatomy features. Our anatomy feature extractor is to detect the landmarks, and then the top of gutta-percha landmark and the curve fitted by the landmarks of tooth apical boundary are used as anatomy features.

\subsubsection{Landmark Detection}
The landmarks uniformly distributed on the apical region of the tooth and the top of the gutta-percha landmark are first detected by HRNet\cite{wang2020deep,sun2019deep}, in which the high-resolution representation is maintained by combining parallel different-resolution representations and repeated multiscale fusion. Among the detected landmarks, the landmark of gutta-percha filled in the root canal is directly used as one of the anatomy features, while the landmarks of tooth apical boundary are fitting to a polynomial curve as the segmentation result, that is, the other significant anatomy feature.

\subsubsection{Fitting Segmentation of Tooth Apical Boundary}
As the most significant anatomy feature, the segmentation of the tooth apical boundary requires very high accuracy. In the traditional segmentation methods, there are many bottlenecks in the judging of the category of each pixel to segment an actual boundary. When segmenting with extremely fuzzy boundaries, not the category for each pixel all can be accurately determined. Our fitting segmentation can be summarized as deriving the actual segmentation boundary according to prior knowledge and determining characteristics.

Our fitting segmentation method is named High-resolution Segmentation based on Polynomial Curve Fitting with Landmark Detection (HS-PCL). In the segmentation of the tooth apical fuzzy boundary, the landmarks in the tooth apical boundary are extracted, and the image is rotated and corrected according to the angle between the root canal and the vertical direction, and we fit the detected landmarks to polynomials of degree $\delta$. We found that the shape characteristics of the apical boundary of the tooth root and the quadratic polynomial curve shown a high degree of similarity. Taking advantage of this, we applied the quadratic polynomial curve fitting of the ordinary least squares method\cite{hutcheson2011ordinary} to the boundary segmentation to bring the segmentation result to be as close as possible to the actual boundary. The quadratic polynomial curve is defined in Eq. (\ref{f1}).

\begin{equation}
\begin{aligned}
y=ax^2 +bx+c\ (a\neq0)
\end{aligned}
\label{f1}
\end{equation}
The matrix equation for the quadratic curve is given in Eq. (\ref{f2}).
According to this, the values of parameters $a$, $b$, and $c$ can be calculated.

\begin{equation}
\begin{aligned}
\left[\begin{matrix}\sum x_i^4&\sum x_i^3&\sum x_i^2\\\sum x_i^3&\sum x_i^2&\sum x_i\\\sum x_i^2&\sum x_i&n\\\end{matrix}\right]\left[\begin{matrix}a\\b\\c\\\end{matrix}\right]=\left[\begin{matrix}\sum{x_i^2y_i}\\\sum{x_iy_i}\\\sum y_i\\\end{matrix}\right]
\end{aligned}
\label{f2}
\end{equation}

where ($x_i$,$y_i$) is the coordinate position of the landmark. Besides, $i$ and $n$ represent the serial number and the total number of the landmarks, respectively. 

Although the landmarks detected by HRNet are not distributed on the boundary with complete accuracy, they can maintain a uniform distribution around the boundary. Therefore, as a segmentation result, the fitting curve can achieve relatively accurate results in the case of uncertain and fuzzy root boundaries. This is shown in \textcolor[rgb]{0,0.541,0.855}{Fig. \ref{fig:osl}}. Our method not only increases the accuracy of the segmentation but also significantly improves the efficiency of the calculation.

\begin{figure*}[ht]
  \centering
  \includegraphics[width=1.\textwidth]{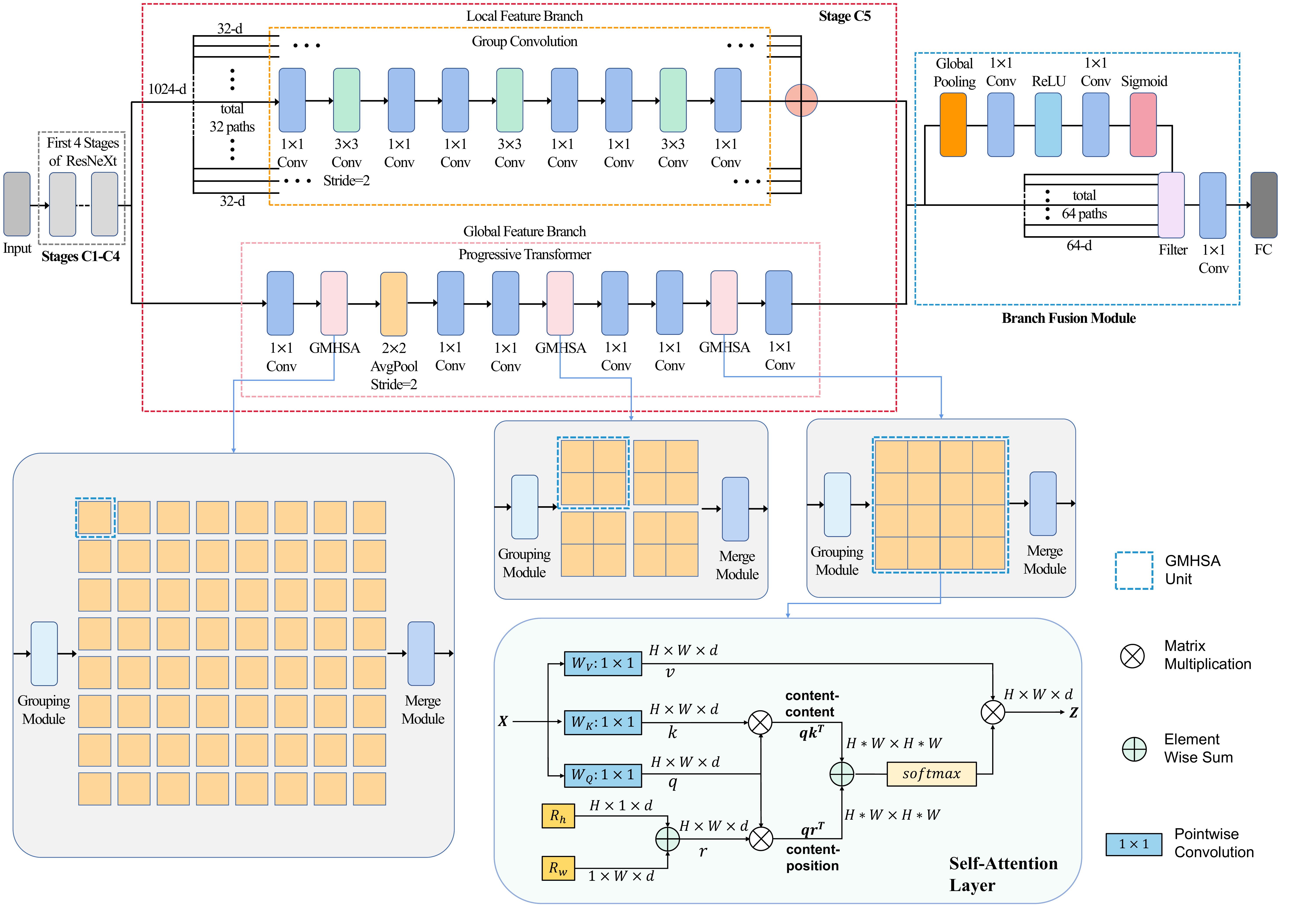}
  \caption{The architecture of our multi-branch Transformer network. The ReLU follows all convolution layers and the skip connection is used between the base block to avoid gradient vanishing. The upper branch is the local feature branch, which is processed by group convolution, and the lower one is the global feature branch, which extracted the global features through progressive Transformer. At the end of the network structure, it is our feature fusion module. Group Multi-Head Self-Attention is embedded in our progressive Transformer. Self-Attention layer is the base structure of GMHSA where our GMHSA has 4 heads, and we do not illustrate them on the figure for simplicity.}
  \label{fig:MBT}
\end{figure*}

\subsection{Multi-Branch Transformer Network for Automatic Evaluation}

While the convolution operation can effectively capture local information, this task may also require global features, such as the shape of the tooth root. In order to globally aggregate the locally acquired features, it is imperative to achieve this through Transformer-based architectures. Therefore, an explicit mechanism for modeling global dependencies could be a more powerful and scalable solution than stacking multiple layers. To combine the advantages of convolution and Transformer, we propose a multi-branch Transformer network, where one is the local feature branch and the other is the global feature branch. An overview of our multi-branch Transformer network is shown in \textcolor[rgb]{0,0.541,0.855}{Fig. \ref{fig:MBT}}.

\subsubsection{Overall Network Structure}
First of all, we will give a brief introduction to our network structure. As shown in \textcolor[rgb]{0,0.541,0.855}{Fig. \ref{fig:MBT}}, the stages C1-C4 of our network adopt the first four stages structure of ResNeXt. In stage C5, our network is divided into a local feature branch and a global feature branch. In the local feature branch, we employ group convolution as the base structure. In the global feature branch, we deploy it based on our progressive Transformer. The final classification result can be obtained through the branch fusion module and the fully connected layer. In this part, we introduced our network as a whole, and in the following paragraphs, we will introduce our network structure in modules.

\subsubsection{Local Feature Branch}
Due to the existence of the multi-branch structure, the local feature branch only needs to focus on better obtaining the local features. As is known to all, convolution extracts the features of different location areas, and the upper layer convolution combines the features of the lower layer convolution and extracts the deeper information. The locality is one of the most important characteristics of convolution. However, ordinary convolution can only realize the locality in space. The proposition of group convolution achieved locality between channels, which further develops the local convolution. In order to obtain local features better, we take the group convolution as the base structure of the local feature branch.

\subsubsection{Global Feature Branch}
In the global feature branch, the ultimate goal is to extract global features such as root shape and X-ray projection angle, etc. To our disappointment, the locality is an inherent characteristic of CNNs, and thus it is necessary to design a new architecture for global feature extraction. Transformer-based architectures use position coding, which can consider the information in different locations simultaneously. Therefore, they can effectively extract global features. Nevertheless, Transformer lacks some of the inductive biases inherent in CNNs, such as locality and translational equivariance, so a hybrid architecture is a suitable solution where CNNs are constructed to extract feature representations and Transformer is constructed to model the remote dependency of the extracted feature maps. Specifically, our progressive Transformer is composed of skip connection, convolutions, average-pooling, and Group Multi-Head Self-Attention.

When a picture is being observed, the same attention cannot be paid everywhere. Hence, different features may be given different importance. A widely applied attention function is scaled dot-product attention, which is calculated in Eq. (\ref{f3}):
\begin{equation}
\begin{aligned}
Attention(Q,K,V)=softmax(QK^T/\sqrt{d_k})V 
\end{aligned}
\label{f3}
\end{equation}
where $d_k$ refers to the dimension of the key. It is known as self-attention when $Q$(queries), $K$(keys), and $V$(values) are equal.
Multi-Head Self-Attention (MHSA)\cite{vaswani2017attention} as the base structure of Transformer is a type of widely used attention mechanism which is more focused on the internal structure. 
Transformer was first proposed for NLP tasks, and so it is difficult to apply Transformer directly to medical image tasks on account of the discrepancy between natural language and image. The number of words in natural language is limited, but the number of pixels increases quadratic with the increase of image size. Because of this, we designed Group Multi-Head Self-Attention in our progressive Transformer to solve the problem of too much computation due to the image characteristics in the medical image. We deploy our progressive Transformer in stage C5 of the network, where our progressive Transformer achieves multi-scale self-attention and reduces the amount of computation. The detailed architecture of our GMHSA is illustrated in \textcolor[rgb]{0,0.541,0.855}{Fig. \ref{fig:MBT}}, where the position encoding method is relative-distance-aware position encoding~\cite{shaw2018self,ramachandran2019stand,Bello_2019_ICCV} with $R_{h}$ and $R_{w}$ for height and width. The attention logit is $qk^T$ + $qr^T$, and $q, k, v, r$ denote query, key, value, and position encodings, respectively. Besides, X represents the feature block obtained after grouping module, and Z represents the output of self attention layer. The grouping module divides the original feature blocks of $C \times H \times W$ into small blocks of $N \times C \times h \times w$, and performs self-attention operation on the N small feature blocks respectively. The merging module merges and restores the feature blocks after grouping operation to the original feature block arrangement.

Assuming that the original feature block size is $H \times W \times C$, and the size of each GMHSA unit is set to $h \times w$, the computation will be greatly reduced through the grouping structure and bottleneck structure. The calculation amount of MHSA before improvement~\cite{liu2021swin} and that of our GMHSA are given in Eq. (\ref{f4}) and (\ref{f5}), respectively:
\begin{equation}
\begin{aligned}
\Omega(MHSA) = 4HWC^2 + 2(HW)^2C
\end{aligned}
\label{f4}
\end{equation}

\begin{equation}
\begin{aligned}
\Omega(GMHSA) &= 4hw(\frac{C}{\varphi})^2 + 2(hw)^2\frac{C}{\varphi} \\
&=\frac{4hw}{\varphi^2HW}HWC^2 + \frac{2h^2w^2}{\varphi H^2W^2}(HW)^2C
\end{aligned}
\label{f5}
\end{equation}
where $\varphi$ is the channel scaling factor of the bottleneck structure, and the size of each GMHSA unit is determined by the input image size. Since GMHSA has not the ability to downsampling, average-pooling with the stride of 2 is implemented for the spatial downsampling following the first GMHSA block. While the feature block size is reduced, the size of GMHSA is gradually increased to achieve multi-scale self-attention and eventually global self-attention.

\begin{figure}[ht]
  \centering
  \includegraphics[width=0.5\textwidth]{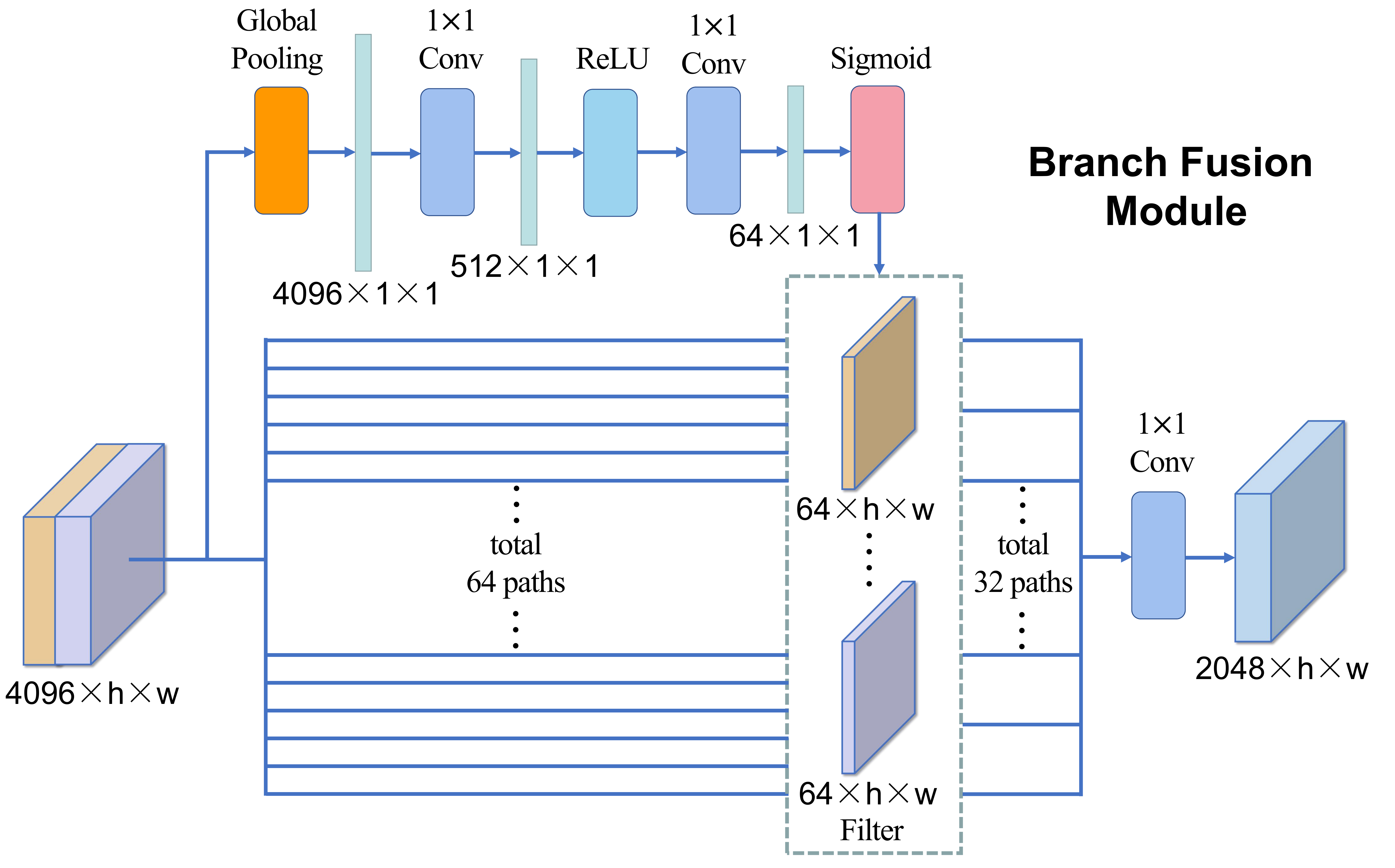}
  \caption{The proposed branch fusion module for two branch feature fusion. The basic structure is to model the channel block interdependencies, remove half of channel block features with lower scores through the channel filter, and fuse the remaining channel features as the output.}
  \label{fig:BFM}
\end{figure}

\subsubsection{Branch Fusion Module}
The multi-branch structure is worth exploring, but it is equally important to better fuse different branch features. The two branches of AGMB-Transformer finally output 2048-channels feature blocks, respectively, and the total number of feature block channels of the two branches is 4096. Too many channels may lead to overfitting and extract more noise features. In order to better fuse the feature blocks of the two branches, we design our branch fusion module following the two branches, improving the quality of branch fusion by explicitly modeling the interdependencies between the channels of two branches' features and filter out half of the channel features for reconnection. Therefore, our branch fusion module can automatically emphasize more informative branch features and suppress less valuable ones. A diagram illustrating the structure of branch fusion module is shown in \textcolor[rgb]{0,0.541,0.855}{Fig. \ref{fig:BFM}}.

Our branch fusion module explicitly models channel interdependencies of two branchs' features to filter the useless features so that the network has the ability to enhance its sensitivity to channel features which can be exploited by the channel filter. In order to generate channel-wise statistics, we squeeze the channel features into a channel descriptor $U$ through global average pooling. Then, two 1×1 convolutions, a ReLU activation, and a sigmoid activation are adopted to obtain the channel block score information for the channel filter. The score $S$ of each channel block is obtained in Eq. (\ref{f6}):
\begin{equation}
\begin{aligned}
S=\sigma (W_{2} \gamma (W_1 U))
\end{aligned}
\label{f6}
\end{equation}

where $\sigma$ denotes sigmoid activation and $\gamma$ is ReLU activation. $W_1$ and $W_2$ belong to the first convolution  $ \mathbb{C}^{4096 \to 512} $ and second convolution $ \mathbb{C}^{512 \to 64} $, respectively. 

After the connection of the two branch features, the 4096 channel features are divided into $\theta$ blocks, and there are $4096/\theta$ channels for each block. According to the channel block scores calculated previously, we filter out the $\theta/2$ blocks with lower scores and reconnect the $\theta/2$ ones with higher scores into a feature block with 2048 channels, and then they are fused by a $1 \times 1$ convolution as the final feature. Moreover, $\theta$ is 64 in this paper.

\section{Experiments and Discussions}
\subsection{Datasets and Experimental Settings}

To evaluate segmentation performance and classification ability, the proposed model was trained and tested on a multi-modality dataset, which includes three kinds of modality annotation: classification, segmentation, and landmarks. In marking the landmarks of the datasets, we marked a total of 19 landmarks from left to right at the tooth root boundary and one landmark of the top of the filled gutta-percha in the root canal. The final dataset consisted of 245 images from different patients, including 90 under-filling images, 96 correct-filling images and 59 over-filling images. These images were collected from the National Clinical Research Center for Oral Diseases, West China Hospital of Stomatology. All the data described in this paper are acquired the ethics approvals (No. 2020484) from Sichuan Academy of Medical Sciences. A stomatologist experienced in oral radiography and root canal therapy classified the images of the tooth to be treated and labeled both the segmentation and the landmarks. Besides, the labeling results were reviewed by two stomatologists to ensure the accuracy of the labeling. When testing the performance both of the segmentation and the segmentation, we adopted 3-fold cross-validation.

\begin{table*}[ht]
\centering
\renewcommand\arraystretch{1.2}
\setlength{\tabcolsep}{5.6mm}
\caption{Comparison of Models}
\begin{tabular}{c|cccccc}
\Xhline{1pt}
Model    & ACC(\%)&AUC(\%) & SEN(\%) & SPC(\%) & F1(\%)   \\ \cline{1-6}

ResNet50  & 57.96$\pm$1.12  & 69.76$\pm$1.17 & 53.36$\pm$4.32  & 77.52$\pm$1.33  & 53.33$\pm$5.17 \\ \cline{1-6}
ResNeXt50  & 56.73$\pm$1.83  & 67.70$\pm$4.07 & 51.80$\pm$4.47  & 76.67$\pm$1.29  & 50.98$\pm$6.07 \\ \cline{1-6}
GCNet50 & 55.93$\pm$2.19  & 63.49$\pm$5.01 & 50.95$\pm$5.87  & 76.21$\pm$2.31  & 50.06$\pm$7.03 \\ \cline{1-6}
BoTNet50  & 56.33$\pm$3.68  & 66.03$\pm$7.50 & 51.32$\pm$6.52  & 76.57$\pm$2.69  & 50.29$\pm$7.88 \\ \cline{1-6}

\textbf{AGMB-Transformer}   & \textbf{90.20$\pm$1.29} & \textbf{95.63$\pm$1.09}    & \textbf{91.39$\pm$1.46}    & \textbf{95.09$\pm$0.77} & \textbf{90.48$\pm$1.13} \\ \Xhline{1pt}
\end{tabular}
\label{tab1}
\end{table*}

For implementation, the models were based on the open-source deep learning framework Pytorch\cite{paszke2019pytorch}. For the compared models, we directly followed the default settings and fine-tuned the baseline model by our best using the official code of the original papers. For optimization technique, we use cosine annealing, which helps the neural network model to converge much faster than using a static learning rate, and it has proved to beats other scheduling techniques in many cases. We use the early stop mechanism in training, which is a widely used method to avoid over fitting, and we set it to trigger early stop when 70 consecutive epoch results are not improved. For each training, 300 training epochs were deployed, with 0.0005 weight decay, a Adam\cite{kingma2014adam} with (0.9, 0.999) betas, 32 cases per minibatch, 0.01 learning rate at the beginning, and the channel scaling factor $\varphi$ is set to $4$. Images were resized to $128 \times 128$ for input. Our data augmentation was based on the fast and flexible image library Albumentations\cite{info11020125}. During the training of all the models, contrast limited adaptive histogram equalization (CLAHE)\cite{reza2004realization} ($clip\ limit=4, p=0.5$) was used to restrain noise signals and enhance the contrast between tissues. Data augmentations including random brightness ($limit=0.1, p=1$), random contrast ($limit=0.1, p=1$), motion blur ($blur\ limit=3, p=0.5$), median blur ($blur\ limit=3, p=0.5$), gaussian blur ($blur\ limit=3, p=0.5$), vertical flip ($p=0.5$) and shift scale rotate ($shift\ limit=0.2, scale\ limit=0.2, rotate\ limit=20, p=1$) were also used to improve robustness and limit the impact of overfitting, where $p$ is the probability value. Core code is available at https://github.com/Kent0n-Li/AGMB-Transformer. 

\subsection{Evaluation Metrics}
We used several metrics for the evaluation of our experiments. Accuracy (ACC), Area Under Curve (AUC), Sensitivity (SEN), Specificity (SPC), and F1 score were employed. 
\begin{itemize}

\item[$\bullet$]Accuracy (ACC):
\begin{equation}
\begin{aligned}
Accuracy = \frac{TP+TN}{TP+TN+FP+FN}
\end{aligned}
\label{f7}
\end{equation}

\item[$\bullet$] Sensitivity (SEN):
\begin{equation}
\begin{aligned}
Sensitivity = \frac{TP}{TP+FN}
\end{aligned}
\label{f8}
\end{equation}

\item[$\bullet$] Specificity (SPC):
\begin{equation}
\begin{aligned}
Specificity = \frac{TN}{TN+FP}
\end{aligned}
\label{f9}
\end{equation}

\item[$\bullet$] F1 score:
\begin{equation}
\begin{aligned}
F1 = \frac{2TP}{2TP+FP+FN}
\end{aligned}
\label{f10}
\end{equation}

\item[$\bullet$] Area Under Curve (AUC):
AUC of the receiver operating character (ROC) is used to compare model functionality.
\end{itemize}
where TP, TN, FP, and FN stand for the number of true positive, true negative, false positive, and false negative predictions. Mean values between classes were calculated to represent the final performance score for multi-class datasets of each model. Specifically, the evaluation of N-classification was divided into N two-classification evaluations, and the metrics of each two classifications were calculated. The average of N metrics is the final result.

We choose the following commonly used metrics in the segmentation of medical images as evaluation criteria to evaluate the segmentation performance of different methods:\\
\textbf{Average Symmetric Surface Distance (ASD):}
\begin{equation}
\begin{aligned}
\frac{( \sum inf_{p \in S_{seg}} d(p,S_{gt})+ \sum inf_{p^{'} \in S_{gt}}d(p^{'},S_{seg}))}{{ \parallel S_{gt} \parallel + \parallel S_{seg} \parallel  }} 
\end{aligned}
\label{f11}
\end{equation}
\textbf{95th-percentile Bidirectional Hausdorff Distance (HD95):}
The Hausdorff distance measures the surface distance between the segmented
and ground truth objects. $d_{H}(S_{seg},S_{gt})$ is defined in Eq. (12):
\begin{equation}
\begin{aligned}
max\{ sup_{a} inf_{b}d(a,b),sup_{b} inf_{a}d(b,a) \},a \in S_{seg}, b \in S_{gt}
\end{aligned}
\label{f12}
\end{equation}

In this paper, we use the 95th-percentile bidirectional Hausdorff distance as a metric for the measure. Where $S_{gt}$ and $S_{seg}$ are the surface voxel sets of manually labeled ground truth and automatically segmented results, respectively, and $d$ is the Euclidean Distance, where $sup$ indicates the upper bound and $inf$ indicates the lower bound of the Euclidean distance between $a$ and $b$.

\begin{table*}[ht]
\centering
\renewcommand\arraystretch{1.2}
\setlength{\tabcolsep}{3mm}
\caption{Evaluation of Segmentation}
\begin{tabular}{c|ccccccc}
\Xhline{1pt}
\textbf{} &
  U-Net &
  \begin{tabular}[c]{@{}c@{}}Attention\\ U-Net\end{tabular} &
  R2U-Net &
  \begin{tabular}[c]{@{}c@{}}Attention\\ R2U-Net\end{tabular} &
  \begin{tabular}[c]{@{}c@{}}U-Net\\ +DFM\end{tabular} &
  HRNet-OCR &
  \textbf{HS-PCL} \\ \hline
ASD(mm)  & 0.547±0.079 & 0.581±0.060 & 0.598±0.193 & 0.538±0.088 & 0.492±0.069 & 0.498±0.090 & \textbf{0.260±0.119} \\
p-value  & 3.2E-7*      & 2.6E-6*      & 8.0E-3*      & 1.3E-5*     & 6.9E-7*      & 1.6E-10*      & N/A                  \\ \hline
HD95(mm) & 1.298±0.062 & 1.343±0.078 & 1.178±0.197 & 1.188±0.112 & 1.085±0.102 & 1.095±0.161 & \textbf{0.763±0.296} \\
p-value  & 1.7E-10*     & 1.2E-12*      & 7.0E-6*      & 8.1E-9*     & 1.1E-6*      & 6.6E-7*     & N/A                  \\ \Xhline{1pt}
\end{tabular}
\label{tab2}
\end{table*}

\subsection{Performance Comparison}
The classification accuracy of different networks is presented in \textcolor[rgb]{0,0.541,0.855}{Table \ref{tab1}}. Our AGMB-Transformer first automatically extracts the anatomy features and then classifies the oral X-ray images with the assistance of the anatomy features, while other networks classify the X-ray images directly. One of the top-performing baseline networks, ResNet50, achieved an accuracy of 57.96\% on the test dataset, but its performance is still far from the goal of the automatic evaluation of root canal therapy, which also means that this task is strongly arduous. Traditional classification networks are not suitable for root canal therapy classification, but our AGMB-Transformer, which extracts anatomy features as the prior knowledge, is an effective method. Considering that the gap between various types is very small, it can be considered as a fine-grained classification task, and the amount of data is very small. Extracting anatomy features as the guide may be an appropriate solution. In addition, AGMB-Transformer pays attention to global features and local features at the same time through the multi-branch structure. This is why our AGMB-Transformer achieves 90.20\% accuracy and has shown such strong improvement.

\subsection{Effectiveness of Each Stage}
Our method composition could be concluded into fitting segmentation, anatomy features, multi-branch Transformer network, and branch fusion module. Parameters were maintained unchanged as far as possible for condition control in all the experiments.

\subsubsection{Effectiveness of Fitting Segmentation}

\begin{figure*}[ht]
\centering
  \includegraphics[width=.85\textwidth]{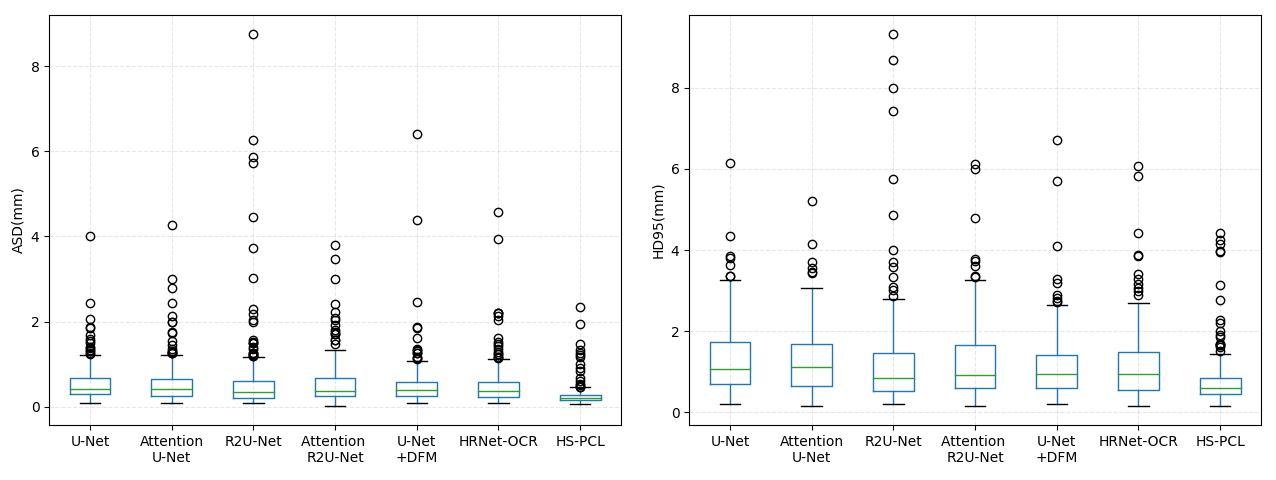}
  \caption{Boxplot of HD95 and ASD values of the tooth apical segmentation with respect to different Methods.}
  \label{fig:asd}
\end{figure*}

\begin{figure*}[ht]
\centering
  \includegraphics[width=1.\textwidth]{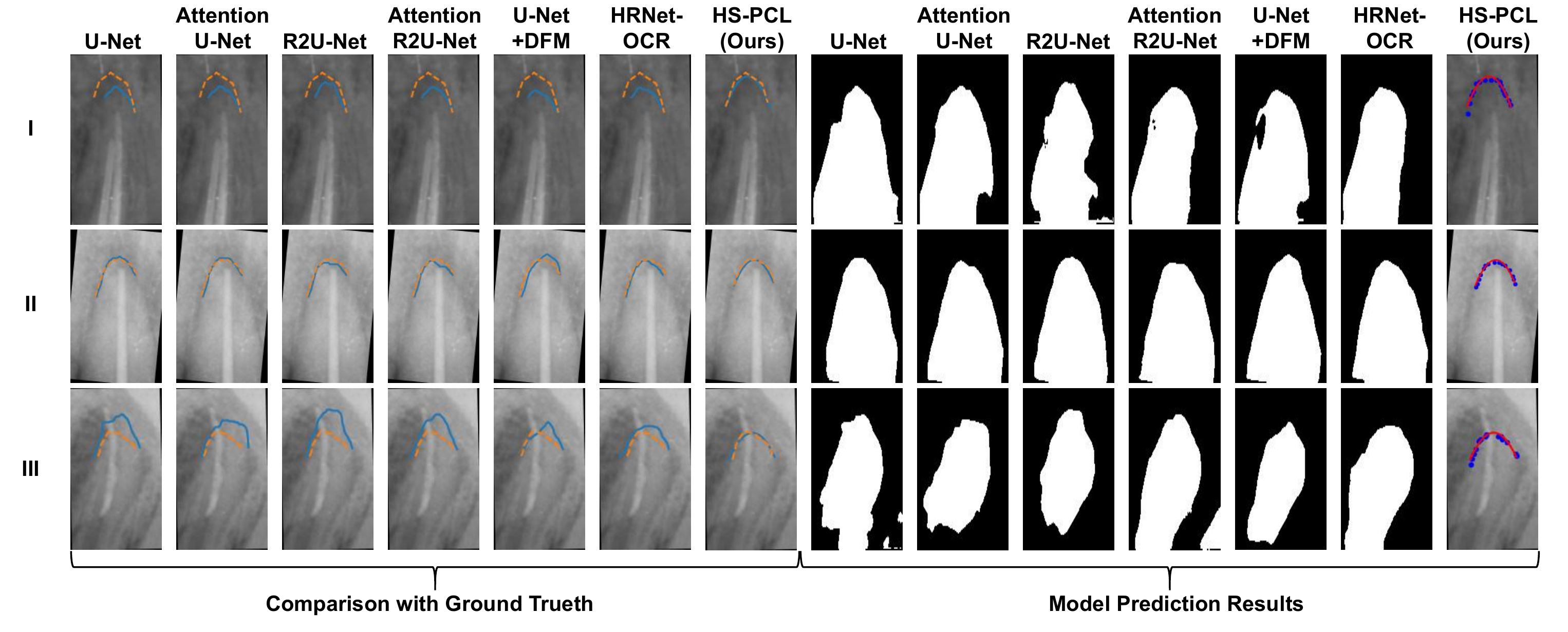}
  \caption{Three examples of the tooth apical area segmentation results of each class. The orange dashed line is the exact position of the tooth apical boundary, and the blue solid line is the model prediction result.}
  \label{fig:segmentation results}
\end{figure*}

\begin{table*}[ht]
\centering
\renewcommand\arraystretch{1.2}
\setlength{\tabcolsep}{6mm}
\caption{Comparison of Different Classification Methods through Anatomy Features}
\begin{tabular}{c|cccccc}
\Xhline{1pt}
Classification Method     & ACC(\%) & SEN(\%) & SPC(\%) & F1(\%)   \\ \hline
Fixed Threshold & 73.33   & 74.69   & 86.93   & 71.19 \\ \hline
\textbf{Classification Network (ResNet50)}   & \textbf{85.30$\pm$1.32}    & \textbf{86.32$\pm$0.80}   & \textbf{92.54$\pm$0.41}   & \textbf{85.74$\pm$1.89}     \\ \Xhline{1pt}
\end{tabular}
\label{tab3}
\end{table*}

The effective position of the segmentation result is the apical area of the tooth root. Therefore, we only marked the landmarks of the apical boundary of the tooth root as the datasets. We rotated the predicted results of baselines as well as the ground truth to the same rotation angle as our model. Additionally, we cropped the same height on the left and right edge to ensure the comparison of the predicted results of each method at the same tooth root position.

\textcolor[rgb]{0,0.541,0.855}{Table \ref{tab2}} makes quantitative comparison of mean ASD and HD95 with a standard deviation with our HS-PCL. Further, we performed a one-tail paired t-test between our method and each comparison method. All the comparing methods whose difference with our method is statistically significant ($p \textless 0.05$) are marked by the symbol * in \textcolor[rgb]{0,0.541,0.855}{Table \ref{tab2}}. The boxplot of ASD and HD95 values of the segmentation with our method and baseline methods is shown in \textcolor[rgb]{0,0.541,0.855}{Fig. \ref{fig:asd}}. It can be seen that our method significantly leads all the baseline methods in both HD95 and ASD metrics. We drew the results predicted by the model with a blue solid line and ground truth with orange dashed line in the visual results as shown in \textcolor[rgb]{0,0.541,0.855}{Fig. \ref{fig:segmentation results}}, which presents a comparison of segmentation results with fuzzy boundaries. We can see from this figure that U-Net and its improved versions performed unsatisfactorily for fuzzy edge segmentation. To verify the effectiveness of our fitting segmentation, we also compared it with HRNet-OCR, which is directly segmented by HRNet. U-Net+DFM, which was proposed for fuzzy boundary segmentation, achieves much better performance than the other networks. Nevertheless, it is still outperformed by our method. 

It is challenging to obtain accurate anatomy features due to the fuzzy boundary of the tooth root. Most traditional deep-learning segmentation methods work by predicting the segmentation map at the pixel level - they assess the category of each pixel, but the boundaries are not always well defined, especially for oral radiographs. In recent years, many researchers have striven to improve U-Net to solve the problem of fuzzy boundary segmentation and have made some progress. Nevertheless, little effort has been made to get rid of the basic architecture based on pixel-level segmentation maps and to perform fuzzy boundary segmentation in a completely different way. Sharp image segmentation can be defined as the partitioning of the pixel set into a family of contiguous subsets or regions. In the segmentation method, there are many bottlenecks in judging the category of each pixel. When segmenting with extremely fuzzy boundaries, not all pixels can have their category accurately determined, and sometimes it is not clear even to experienced dentists which category each pixel should belong to. When dentists segment the fuzzy boundary of the tooth root, they often combine it with the anatomical context and some determinable features. This can be summarized as deriving the actual segmentation boundary according to prior knowledge, which has proven to be very effective for the segmentation of extremely fuzzy boundaries.

\begin{table*}[ht]
\centering
\renewcommand\arraystretch{1.2}
\setlength{\tabcolsep}{5mm}
\caption{Comparison of Different Inputs on ResNet50}
\begin{tabular}{c|cccccc}
\Xhline{1pt}
Input     & ACC(\%)&AUC(\%) & SEN(\%) & SPC(\%) & F1(\%)   \\ \hline
Image    & 57.96$\pm$1.12  & 69.76$\pm$1.17 & 53.36$\pm$4.32  & 77.52$\pm$1.33  & 53.33$\pm$5.17 \\ \hline
Anatomy Feature & 85.30$\pm$1.32 & 93.82$\pm$1.76   & 86.32$\pm$0.80   & 92.54$\pm$0.41   & 85.74$\pm$1.89 \\ \hline
\textbf{Image+Anatomy Feature}   & \textbf{88.16$\pm$1.50}      & \textbf{94.90$\pm$0.49}       & \textbf{89.27$\pm$2.62}     & \textbf{94.00$\pm$1.01}    & \textbf{88.56$\pm$1.73}      \\ \Xhline{1pt}
\end{tabular}
\label{tab4}
\end{table*}

\begin{table*}[ht]
\centering
\renewcommand\arraystretch{1.2}
\caption{Evaluation of Classification Models with Anatomy Features}
\begin{tabular}{c|ccccc|ccc}
\Xhline{1pt}
\multirow{2}{*}{Model} & \multirow{2}{*}{ACC(\%)} & \multirow{2}{*}{AUC(\%)} & \multirow{2}{*}{SEN(\%)} & \multirow{2}{*}{SPC(\%)} & \multirow{2}{*}{F1(\%)} & \multicolumn{3}{c}{t-test p-value} \\ \cline{7-9} 
                       &                          &                          &                          &                          &                         &  U        & C      & O    \\ \hline

ResNet50               & 88.16$\pm$1.50      & 94.90$\pm$0.49       & 89.27$\pm$2.62     & 94.00$\pm$1.01    & 88.56$\pm$1.73     & 1.6E-5*             & 1.4E-14* & 1.6E-12*           \\\cline{1-9}

ResNeXt50              & 86.94$\pm$1.37       & 95.02$\pm$0.87      & 87.67$\pm$1.14        & 93.20$\pm$0.31     & 87.64$\pm$1.77   & 7.9E-4*    & 1.5E-21*  & 6.9E-9*         \\\cline{1-9}
Xception            & 88.97$\pm$2.16      & 95.04$\pm$0.69    & 89.83$\pm$1.62    & 94.37$\pm$1.26      & 89.41$\pm$1.56    & 1.1E-6*    & 4.1E-6*   &   8.0E-7*        \\ \cline{1-9}

DenseNet169  & 87.75$\pm$1.30   & 94.80$\pm$0.51   & 89.13$\pm$1.50  & 93.85$\pm$0.77   & 88.12$\pm$1.25  & 3.0E-1     & 3.7E-4*   &  4.1E-6*              \\ \cline{1-9}
SEResNet50  & 87.74$\pm$2.53 & 94.38$\pm$1.20    & 88.96$\pm$2.98    & 93.76$\pm$1.24    & 88.13$\pm$2.77    & 2.1E-1     & 1.7E-8*    &   2.3E-8*         \\\cline{1-9}
SEResNeXt50  & 88.98$\pm$1.22  & 94.65$\pm$2.20    & 90.44$\pm$1.06  & 94.35$\pm$0.73     & 89.56$\pm$0.98  & 4.1E-2*  & 1.9E-1  &   8.6E-18*   \\ \cline{1-9}
GCNet50  & 88.56$\pm$1.95     & 94.65$\pm$0.64    & 90.17$\pm$1.97      & 94.34$\pm$0.87  & 88.91$\pm$2.09     & 1.1E-5*         & 2.4E-3*              &  8.6E-5*           \\ \cline{1-9}
BoTNet   & 89.38$\pm$1.93        & 94.81$\pm$0.42    & 89.97$\pm$2.25 & 94.62$\pm$1.11   & 89.51$\pm$1.77   & 2.1E-2*    & 1.5E-2*      & 1.9E-17*       \\ \cline{1-9}
ViT & 60.44$\pm$6.49   & 72.31$\pm$4.85     & 59.67$\pm$8.10   & 78.89$\pm$4.15   & 59.70$\pm$7.48     & 4.0E-2*   & 3.6E-2*      & 7.1E-10*     \\ \cline{1-9}

\textbf{AGMB-Transformer}   & \textbf{90.20$\pm$1.29} & \textbf{95.63$\pm$1.09}    & \textbf{91.39$\pm$1.46} & \textbf{95.09$\pm$0.77} & \textbf{90.48$\pm$1.13}  & N/A  & N/A  & N/A \\ \Xhline{1pt}
\end{tabular}
\label{tab5}
\end{table*}

\begin{figure*}[ht]
\centering
  \includegraphics[width=.85\textwidth]{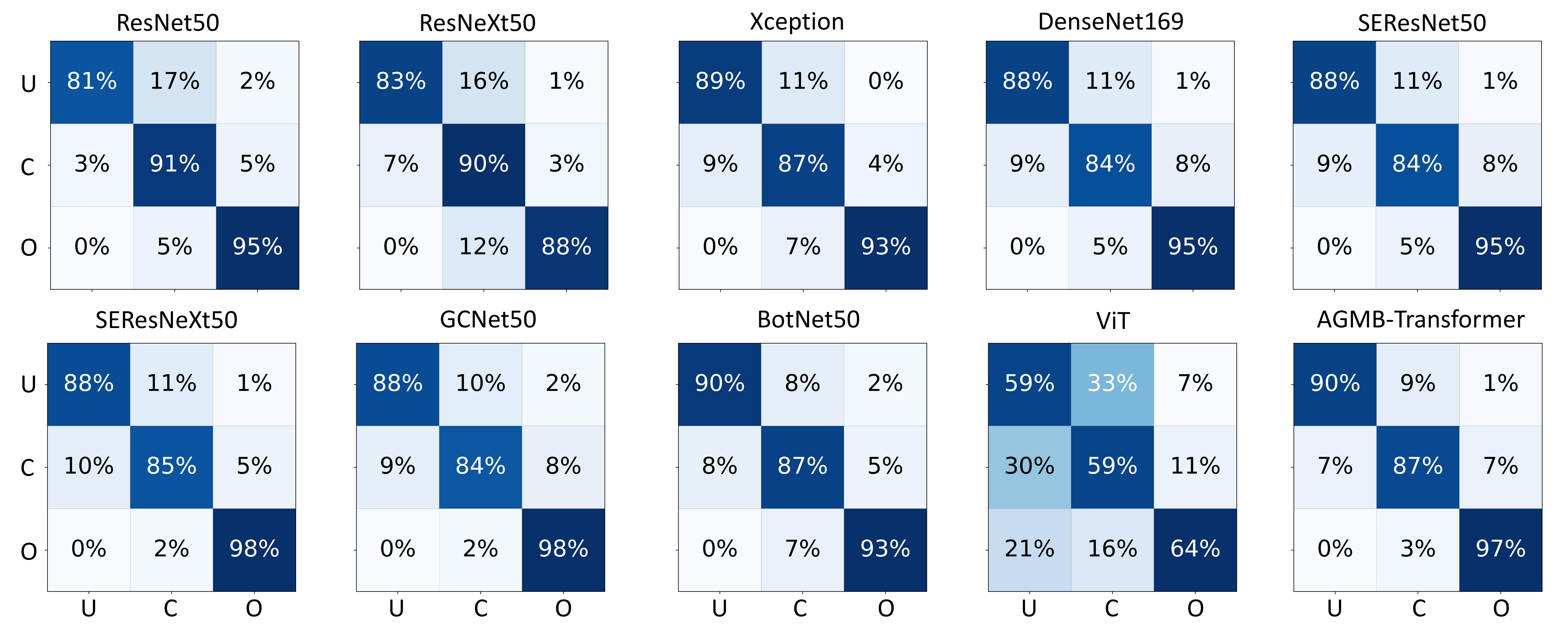}
  \caption{Ten charts of confusion matrices demonstrate the distribution of predictions. The color of confusion matrices provides better visualization, and the depth of the color depends on the normalized values of predictions. Symbols used in the figure are denoted as U: Under-filling, C: Correct-filling, O: Over-filling.}
  \label{fig:matrix}
\end{figure*}

\subsubsection{Effectiveness of Anatomy Feature}
Normally, root shape and X-ray projection angle are also necessary for the evaluation of root canal therapy. Models trained by X-ray images with anatomy features and classifying by threshold by measuring the distance directly are compared in \textcolor[rgb]{0,0.541,0.855}{Table \ref{tab3}}. In this circumstance, it is not accurate to measure the distance between the apical area boundary of the tooth root and the top of the filled gutta-percha. 

In our implementation, we designed an anatomy-guided network to extract the tooth apical boundary and the landmark of gutta-percha in the root canal as prior knowledge and then guided the classification network. To evaluate its importance and effectiveness, we defined two more inputs for comparison. The first input is the 3-channel X-ray image alone. The other has the anatomy feature as the input with only one channel. \textcolor[rgb]{0,0.541,0.855}{Table \ref{tab4}} reports the classification performance based on three different inputs on ResNet50. From these results, we can see that anatomy features greatly enhance classification performance, which demonstrates the guidance capability of boundary and landmark information for diagnosis. When compared with putting them together, having the anatomy feature as the only input is not very effective, which proves that tooth root shape and X-ray projection angle in the original image is also necessary. Additionally, ResNet50 with the anatomy feature and X-ray images as input achieves the best diagnosis performance, revealing that classification network, anatomy features, and original images are all indispensable.

\subsubsection{Effectiveness of Multi-Branch Transformer Network}

The advantages of architecture design have been deeply explored. This has been done by evaluating classic deep neural structures (i.e, ResNet50, ResNeXt50, Xception and DenseNet169), widely used fine-grained image classification networks (i.e, SEResNet, SEResNeXt), high performance non-local networks (i.e, GCNet50), and models with Transformer (i.e, BoTNet, ViT).  To evaluate the importance and effectiveness of AGMB-Transformer, we did a quantitative comparison between our method and the baseline networks on the same dataset using the mean ACC, AUC, SEN, SPC, and F1 with standard deviations. This is shown in \textcolor[rgb]{0,0.541,0.855}{Table \ref{tab5}}. To evaluate the response level of the model to the proposed improved structures, we perform a paired t-test to test the statistical significance of the model’s output. Set t-test be calculated as $p$-value = t-test $(N1, N2)$, where t = $\frac{ \overline{x}- \mu_0}{\frac{s }{ \sqrt{n} }} $, where $\mu_0$ = the test value, $\overline{x}$ = sample mean, $n$ = sample size, $s$ = sample standard deviation, and $N1$ and $N2$ are the two distribution to be evaluated. The p-value can be found from the t-Distribution Table or be automatically calculated by statistical program (R, SPSS, etc.).
We collected the output distributions of all comparison models and our model according to under-filling, correct-filling, and over-filling, respectively: $(U1, U2, ...), (C1, C2, ...), (O1, O2, ...)$. T-test uses t-distribution theory to infer the probability of difference, which is the most common significance test. For each category, we performed a significance test with t-test to assess the extent of the effect on the improved model by every two models’ output response, and splitting the statistical analysis for different types is more helpful to explain that the difference of output results of different networks is significant.
All comparing methods, which showed a difference from our method that was statistically significant ($p \textless 0.05$) are marked by symbol * in \textcolor[rgb]{0,0.541,0.855}{Table \ref{tab5}} ($U$: Under-filling, $C$: Correct-filling, $O$: Over-filling).

\begin{table*}[ht]
\centering
\renewcommand\arraystretch{1.2}
\caption{Comparison of Different Network Components}
\begin{tabular}{c|cccccc}
\Xhline{1pt}
Component   & ACC(\%)&AUC(\%) & SEN(\%) & SPC(\%) & F1(\%)   \\ \hline
Local Feature Branch & 86.94$\pm$1.37   & 95.02$\pm$0.87   & 87.67$\pm$1.14   & 93.20$\pm$0.31   & 87.64$\pm$1.77   \\ \hline
Global Feature Branch & 88.98$\pm$0.08   & 95.38$\pm$0.70   & 90.55$\pm$0.23   & 94.57$\pm$0.18   & 89.34$\pm$0.02   \\ \hline
Global Feature Branch + Local Feature Branch & 89.38$\pm$1.93  & 95.00$\pm$0.55 & 90.57$\pm$2.09  & 94.64$\pm$1.05  & 89.72$\pm$1.89 \\ \hline
\textbf{Local Feature Branch+Global Feature Branch+Branch Fusion Module}      & \textbf{90.20$\pm$1.29} & \textbf{95.63$\pm$1.09}    & \textbf{91.39$\pm$1.46} & \textbf{95.09$\pm$0.77} & \textbf{90.48$\pm$1.13}     \\ \Xhline{1pt}
\end{tabular}
\label{tab6}
\end{table*}

All the networks were trained and tested with the assistance of anatomy features. 
From the experimental results, we can see that, all the networks have improved a lot with the assistance of anatomy features. When compared with other networks, our AGMB-Transformer achieves the best results. Due to the structure of multiple branches, our model achieves self-attention through the global feature branch and retains the advantage of locality group convolution. This is why our model achieves a better result than all others. As we know, the amount of training data required for ViT is 14-300 million images, and so we loaded the pre-trained weight obtained from the training on imagenet1k for ViT. However, the performance of ViT in this task is still very poor. Compared with other models, we can observe that our method has at least two categories with a statistically significant improvement.

To further evaluate the performance of the algorithm, it was necessary to report the distribution of our prediction result, which is shown by the confusion matrix in \textcolor[rgb]{0,0.541,0.855}{Fig. \ref{fig:matrix}}. We can infer that our model is better than all others for discriminating under-filling, as our AGMB-Transformer achieved the highest accuracy of all comparison methods in correctly identifying under-filling. One possible explanation is that the global feature branch models long-range dependencies. The convolution unit in CNNs only pays attention to the range of kernel size in the neighborhood. Even if the receptive field becomes larger and larger in the later stage, it is still a local-area operation that ignores the contribution of other global areas, such as distant pixels, to the current area. This is why classification networks based on CNNs cannot perform the task effectively, but the progressive Transformer shows great improvement.

\subsubsection{Effectiveness of Global Feature Branch and Branch Fusion Module}
To further verify the respective effects of the global feature branch and the branch fusion module in a multi-branch Transformer network, we test the performance of removing the global feature branch and removing the branch fusion module. A comparison with the performance of removing them shows that our global feature branch and our branch fusion module can significantly improve the performance of the network. As the result shown in \textcolor[rgb]{0,0.541,0.855}{Table \ref{tab6}}, all components of multi-branch Transformer network are indispensable.

\subsection{Model Size, Computational Cost and GPU Memory Usage}
Due to the difference between NLP task and computer vision task, direct application of Transformer in computer vision task has the problem of too much computational cost and GPU memory usage. We will compare the effectiveness of our method from three aspects: model size, computational cost and GPU memory usage.

\begin{table}[ht]
\centering
\renewcommand\arraystretch{1.2}
\setlength{\tabcolsep}{0.6mm}
\caption{Comparison on Model Size}
\begin{tabular}{c|cccccc}
\Xhline{1pt}
Model           & ResNet50 & ResNeXt50 & GCNet50 & BoTNet50 & ViT   & Ours  \\ \hline
Size(MB) & 188    & 176     & 286   & 151    & 689 & 358 \\ \Xhline{1pt}
\end{tabular}
\label{tabpara}
\end{table}

As we know, computational cost and GPU memory usage of MHSA will increase sharply with the increase of image size. Therefore, we tested the computational cost and GPU memory usage of our GMHSA and MHSA when inputting feature blocks of different sizes, and we set the size of GMHSA unit to $8 \times 8$ during the experiment.

As shown in \textcolor[rgb]{0,0.541,0.855}{Table \ref{tabpara}}, our model size has not increased very much and is far lower than ViT. Moreover, the performance of our GMHSA is much better than MHSA in more important computational cost and GPU memory usage. Considering that a $C\times H\times W$ feature block will produce the matrix of $ H\times W \times H\times W$ size in the operation of the self-attention layer, which can be seen from \textcolor[rgb]{0,0.541,0.855}{Fig. \ref{fig:MBT}}, and so our GMHSA can better deal with large feature blocks. We can see from \textcolor[rgb]{0,0.541,0.855}{Table \ref{tabflops}} that the $16 \times 144 \times 144$ size feature block needs about 41G GPU memory and 14G FLOPs with MHSA. While with GMHSA, it only needs 1.6G GPU memory and 0.068G FLOPs.

In order to test the effect of our bottleneck structure, we tested the computational cost and GPU memory usage of our GMHSA under the input of feature blocks with different channel amounts. The channel scaling factor $\varphi$ of our bottleneck structure is set to 4. As shown in Tabel \ref{tabchannel}, FLOPs have been decreased from 102.54G to 10.53G when the original channel is 4096.

\begin{table}[ht]
\centering
\renewcommand\arraystretch{1.2}
\setlength{\tabcolsep}{0.6mm}
\caption{Comparison of MHSA and GMHSA}
\begin{tabular}{c|cc|cc}
\Xhline{1pt}
& \multicolumn{2}{c|}{MHSA} & \multicolumn{2}{c}{GMHSA} \\ \hline
Feature Size & Memory(MB)  & FLOPs(G)  & Memory(MB)  & FLOPs(G)         \\ \hline
16 $\times$ 40 $\times$ 40   & 1699      & 0.087    & 1461      & 0.005     \\
16 $\times$ 56 $\times$ 56  & 2373      & 0.328     & 1465      & 0.010      \\
16 $\times$ 72 $\times$ 72   & 3951      & 0.888     & 1487      & 0.017      \\
16 $\times$ 88 $\times$ 88   & 7007      & 1.968    & 1511      & 0.026     \\
16 $\times$ 104 $\times$ 104 & 12273     & 3.824     & 1535      & 0.036      \\
16 $\times$ 120$\times$ 120   & 20611     & 6.757    & 1563      & 0.047     \\
16 $\times$ 136 $\times$ 136   & 33025     & 11.125    & 1579      & 0.061      \\
16 $\times$ 144  $\times$ 144     & 41107     & 13.966     & 1611      & 0.068     \\\Xhline{1pt}
\end{tabular}
\label{tabflops}
\end{table}

\begin{table}[ht]
\centering
\renewcommand\arraystretch{1.2}
\setlength{\tabcolsep}{2.6mm}
\caption{Comparison on Different Feature Channels for GMHSA}
\begin{tabular}{c|cc}
\Xhline{1pt}
Feature Size & Memory(MB) & FLOPs(G) \\ \hline
128$\times$ 40 $\times$ 40          & 1713       & 0.77 \\
256 $\times$ 40 $\times$ 40         & 1739       & 1.69  \\
512 $\times$ 40 $\times$ 40         & 1765       & 4.01  \\
1024 $\times$ 40 $\times$ 40        & 1819       & 10.53  \\
2048 $\times$ 40 $\times$ 40        & 1983       & 31.14  \\
4096 $\times$ 40 $\times$ 40        & 2475       & 102.54  \\\Xhline{1pt}
\end{tabular}
\label{tabchannel}
\end{table}

Through the above results, we can determine that our progressive Transformer, whether GMHSA or bottleneck structure, can well reduce the computational cost and GPU memory usage, which is of great significance for the application of Transformer in computer vision tasks.

\subsection{Impact of Parameters $\theta$ and Degree of Polynomial $\delta$}
There are two other parameters that impact the network performance, namely, $\theta$ in branch fusion module and degree of polynomial $\delta$. To evaluate the sensitivity of performance to them, we change $\theta$ and $\delta$ separately and report the classification performance in \textcolor[rgb]{0,0.541,0.855}{Table \ref{tab:theta}} and \textcolor[rgb]{0,0.541,0.855}{Table \ref{tab:degree}}. From \textcolor[rgb]{0,0.541,0.855}{Table \ref{tab:theta}}, we can see that the smaller or larger $\theta$ values will lead to too big or too small feature blocks, which cannot filter feature blocks well. When $\theta$ = 64, the 4096 channel features are divided into 64 blocks, and there are 64 channels for each block, and the performance is the best. When the degree of polynomial $\delta$ gradually increases from 2 to 5 in \textcolor[rgb]{0,0.541,0.855}{Table \ref{tab:degree}}, the performance of the model will drop. Therefore, we set $\theta$ = 64 and $\delta$ = 2 in our implementation.

\begin{table}[ht]
\centering
\renewcommand\arraystretch{1.2}
\setlength{\tabcolsep}{1.0mm}
\caption{Impact of Parameter $\theta$}
\begin{tabular}{c|cccccc}
\Xhline{1pt}
$\theta$  & ACC(\%)&AUC(\%) & SEN(\%) & SPC(\%) & F1(\%)   \\ \cline{1-6}
32  & 88.97$\pm$1.30  & 95.27$\pm$0.62  & 89.88$\pm$0.60  & 94.50$\pm$0.56  & 88.85$\pm$0.86 \\ 
\textbf{64}   & \textbf{90.20$\pm$1.29} & \textbf{95.63$\pm$1.09}    & \textbf{91.39$\pm$1.46} & \textbf{95.09$\pm$0.77} & \textbf{90.48$\pm$1.13}  \\
128 & 88.16$\pm$1.94  & 94.73$\pm$1.61 & 88.95$\pm$2.27  & 94.02$\pm$1.26 & 88.23$\pm$1.56  \\ 
256 & 88.97$\pm$1.30  & 95.44$\pm$0.57 &  90.04$\pm$0.99  & 94.47$\pm$0.64 & 89.29$\pm$1.12 \\ \Xhline{1pt}
\end{tabular}
\label{tab:theta}
\end{table}

\begin{table}[ht]
\centering
\renewcommand\arraystretch{1.2}
\setlength{\tabcolsep}{3.0mm}
\caption{Impact of Parameter $\delta$}
\begin{tabular}{c|cc}
\Xhline{1pt}
Degree $\delta$ & ASD (mm) & HD95 (mm) \\ \hline
\textbf{2}      & \textbf{0.260$\pm$0.119 }   & \textbf{0.763$\pm$0.296 }    \\
3      & 0.308$\pm$0.114    & 0.885$\pm$0.313     \\
4      & 0.336$\pm$0.032     & 1.029$\pm$0.104   \\
5      & 0.288$\pm$0.100    & 0.884$\pm$0.276  \\\Xhline{1pt}
\end{tabular}
\label{tab:degree}
\end{table}

\section{Conclusion}
In this paper, we have proposed an AGMB-Transformer that is state-of-the-art in automatic X-ray image diagnosis of root canal therapy. The proposed novel method first extracts the anatomy features and then uses them to guide a multi-branch Transformer network for evaluation, where progressive Transformer is an explicit mechanism to model global dependencies. Besides, the fitting segmentation in our anatomy feature extractor gets rid of the basic architecture based on pixel-level segmentation maps and performs fuzzy boundary segmentation in a completely different way. Combining the multiple network structures in parallel is worth exploring since the features obtained by different network structures have their own advantages. Nevertheless, it is a challenge to better fuse multiple network features. We designed a multi-branch structure where our branch fusion module can be an effective solution. Moreover, it is convenient to combine our multi-branch structure and branch fusion module with other networks. In theory, this can be extended to almost all existing backbone networks, which indicates a strong potential for its application to most existing medical imaging pattern recognition models. 

To the best of our knowledge, this is the first time automatic root canal therapy evaluation has been realized utilizing the anatomy feature as prior knowledge. 
However, there are still some limitations in the current work.
In our future work, we will further improve diagnosis performance from the following two aspects: 1) to simplify the operation, we can investigate in a more effective, end-to-end fashion, instead of in the current multi-step one; 2) we can further explore the performance and combined effect of the multi-branch structure and various state-of-the-art networks.

\section*{Appendix}
The performance test of classification network often needs a lot of data, while the number of root canal therapy images used in this paper is too small to exactly compare the performance of the network. To further evaluate the performance of our AGMB-Transformer, we applied it on the widely used public Plant Pathology 2020 challenge dataset\cite{thapa2020plant}. We divided 1821 labeled images into 60\% training set, 20\% validation set and 20\% test set to test our network. The dataset consists of 4 categories leaves: healthy, apple rust, apple scab, and more than one disease. We resize the images to $256 \times 256$, and other parameters are exactly the same as those in the root canal therapy dataset. The comparison results are reported in \textcolor[rgb]{0,0.541,0.855}{Table~\ref{tabpubdata}}. It can be observed that our AGMB-Transformer outperforms other methods on ACC, AUC, SPC and F1.

\begin{table}[ht]
\centering
\renewcommand\arraystretch{1.2}
\setlength{\tabcolsep}{1.0mm}
\caption{Comparison on Public Datasets}
\begin{tabular}{c|cccccc}
\Xhline{1pt}
Model    & ACC(\%)&AUC(\%) & SEN(\%) & SPC(\%) & F1(\%)   \\ \cline{1-6}

ResNet50  & 93.13  & 89.36 & 75.07  & 97.51 & 74.05  \\ \cline{1-6}
ResNeXt50  & 95.05  &89.21   & 81.57  & 98.14  &84.19  \\ \cline{1-6}
GCNet50 & 93.96  & 90.50 &  \textbf{83.96}  & 97.91  & 85.44  \\ \cline{1-6}
BoTNet50  & 93.68  & 89.17  & 74.36  & 97.64  & 72.27  \\ \cline{1-6}

\textbf{AGMB-Transformer}   & \textbf{96.15 } & \textbf{90.56}     & 81.29    & \textbf{99.13 } & \textbf{85.96}  \\ \Xhline{1pt}
\end{tabular}
\label{tabpubdata}
\end{table}

\bibliographystyle{ieeetr}
\bibliography{paper}

\end{document}